%% file: main.tex
\documentclass[10pt,twocolumn,letterpaper]{article}

\usepackage{cvpr}              % To produce the CAMERA-READY version
\input{preamble}

\definecolor{cvprblue}{rgb}{0.21,0.49,0.74}
\usepackage[pagebackref,breaklinks,colorlinks,allcolors=cvprblue]{hyperref}

\def\paperID{}
\def\confName{3D-LLM/VLA Workshop}
\def\confYear{2026}

\title{Reference-Free Assessment of Physical Consistency \\
in World Model-based Video Generation}

\author{
    Yun Oh$^1$
    \qquad Sukmin Yun$^1$
    \thanks{Correspondence to: \href{mailto:sukminyun@hanyang.ac.kr}{sukminyun@hanyang.ac.kr}}
}

\begin{document}
\maketitle

\footnotetext[1]{Hanyang University ERICA}
\input{sec/0_abstract}
\input{sec/1_intro}
\input{sec/2_method}
\input{sec/3_exp}
\input{sec/4_discussion}
{
    \small
    \bibliographystyle{ieeenat_fullname}
    \bibliography{main}
}


\end{document}

%% file: preamble.tex
\usepackage{makecell}
\usepackage{multirow}
\usepackage{stfloats}
\usepackage[table]{xcolor}
\usepackage{graphicx}
\usepackage{float}
\usepackage[T1]{fontenc}

%% file: sec/0_abstract.tex
\begin{abstract}
We introduce reference-free measures for evaluating the physical consistency of generated videos, combining relative and absolute approaches to assess fidelity. Although tools like WorldGym or WorldEval enable robotic simulation via video generation, physical fidelity gaps often prevent these environments from accurately reproducing real-world task success rates of VLA models. Unlike existing evaluation methods, which require costly human voting (Elo) or unavailable ground-truth references (FVD), our approach utilizes DROID-SLAM and SEA-RAFT to quantify physical inconsistencies, motivated by WorldScore. Videos filtered using our relative consistency assessment show an improvement in task success rates of over 8\%, effectively narrowing the simulation-to-reality gap. Furthermore, our absolute assessment enables spatio-temporal localization, providing visualization of when and where physical artifacts occur.
\end{abstract}

%% file: sec/1_intro.tex
\section{Introduction}
\label{sec:intro}

In robot learning, simulation has become a practical necessity due to the limited accessibility of real-world robotic systems. However, performance in simulation often fails to reliably transfer to real-world settings.
Recent VLA models \cite{kim2025finetuningvisionlanguageactionmodelsoptimizing, zheng2025xvlasoftpromptedtransformerscalable, black2026pi0visionlanguageactionflowmodel, kim2026cosmospolicyfinetuningvideo} show task success rate above 95\% on the LIBERO \cite{liu2023liberobenchmarkingknowledgetransfer} benchmark. In contrast, more recent environments \cite{han2025robocerebralargescalebenchmarklonghorizon, zhang2024vlabenchlargescalebenchmarklanguageconditioned} are sufficiently difficult that many existing policies struggle to achieve meaningful success rates.

Meanwhile, world model–based approaches \cite{quevedo2025worldgymworldmodelenvironment, li2025worldevalworldmodelrealworld} have shown promising alignment with real-world outcomes, suggesting a potential path toward more realistic evaluation. Nevertheless, these approaches introduce a new challenge: the generated visual rollouts often suffer from physical inconsistencies, such as object morphing, making it difficult to reliably assess task success. Standard metrics for videos like FVD \cite{unterthiner2019accurategenerativemodelsvideo} fall short here, as they measure statistical distribution rather than the pixel-level physical integrity required for robotic manipulation.

\begin{figure}
  \centering
  \includegraphics[width=0.9\columnwidth]{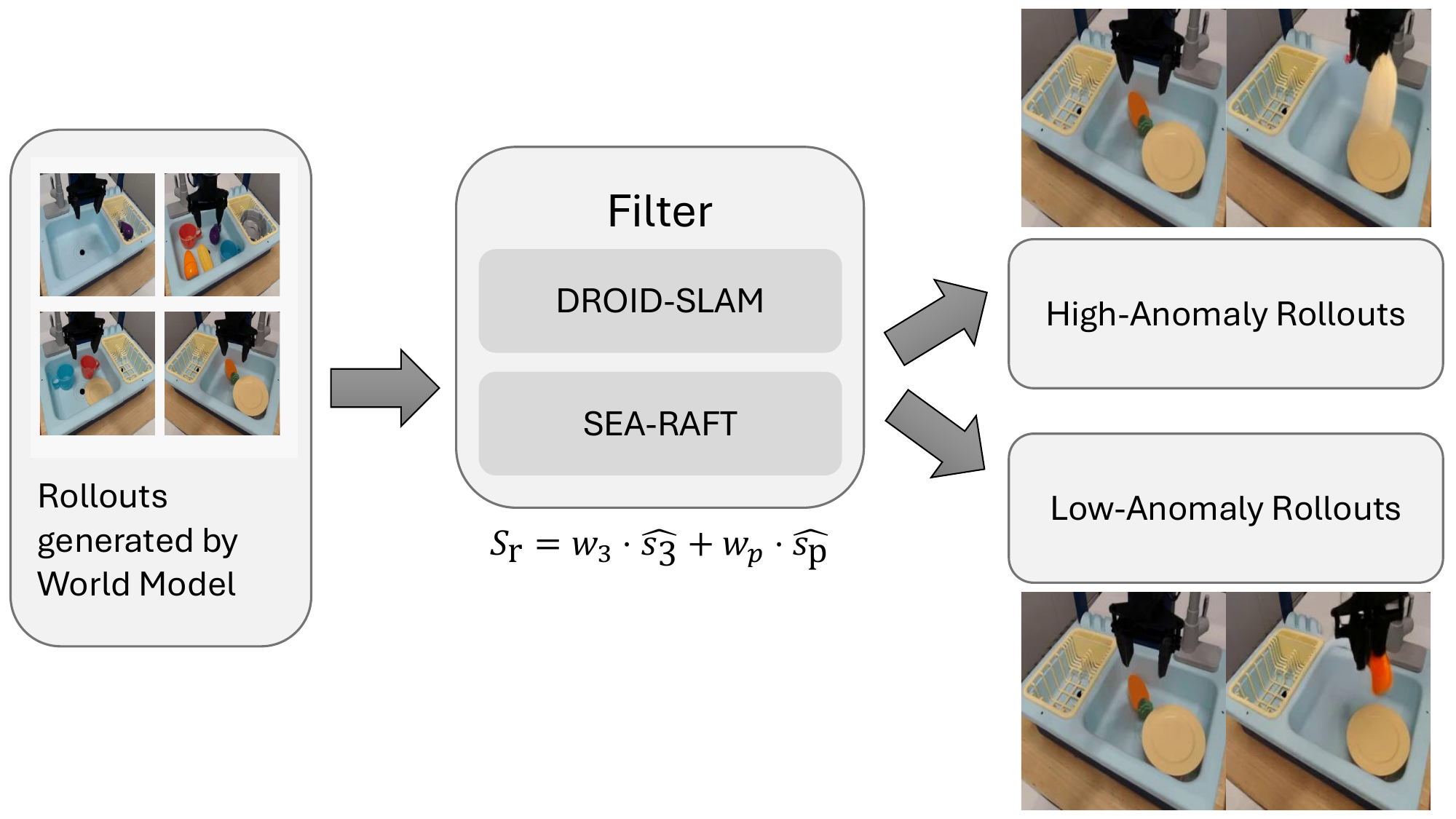}
  \caption{\textbf{Overview of Relative Assessment}. Generated rollouts are evaluated using DROID-SLAM \cite{teed2022droidslamdeepvisualslam} and SEA-RAFT \cite{wang2024searaftsimpleefficientaccurate} to obtain relative anomaly scores, which enable filtering and ranking of samples based on physical plausibility.}
\end{figure}
\begin{figure}
  \centering
  \includegraphics[width=0.9\columnwidth]{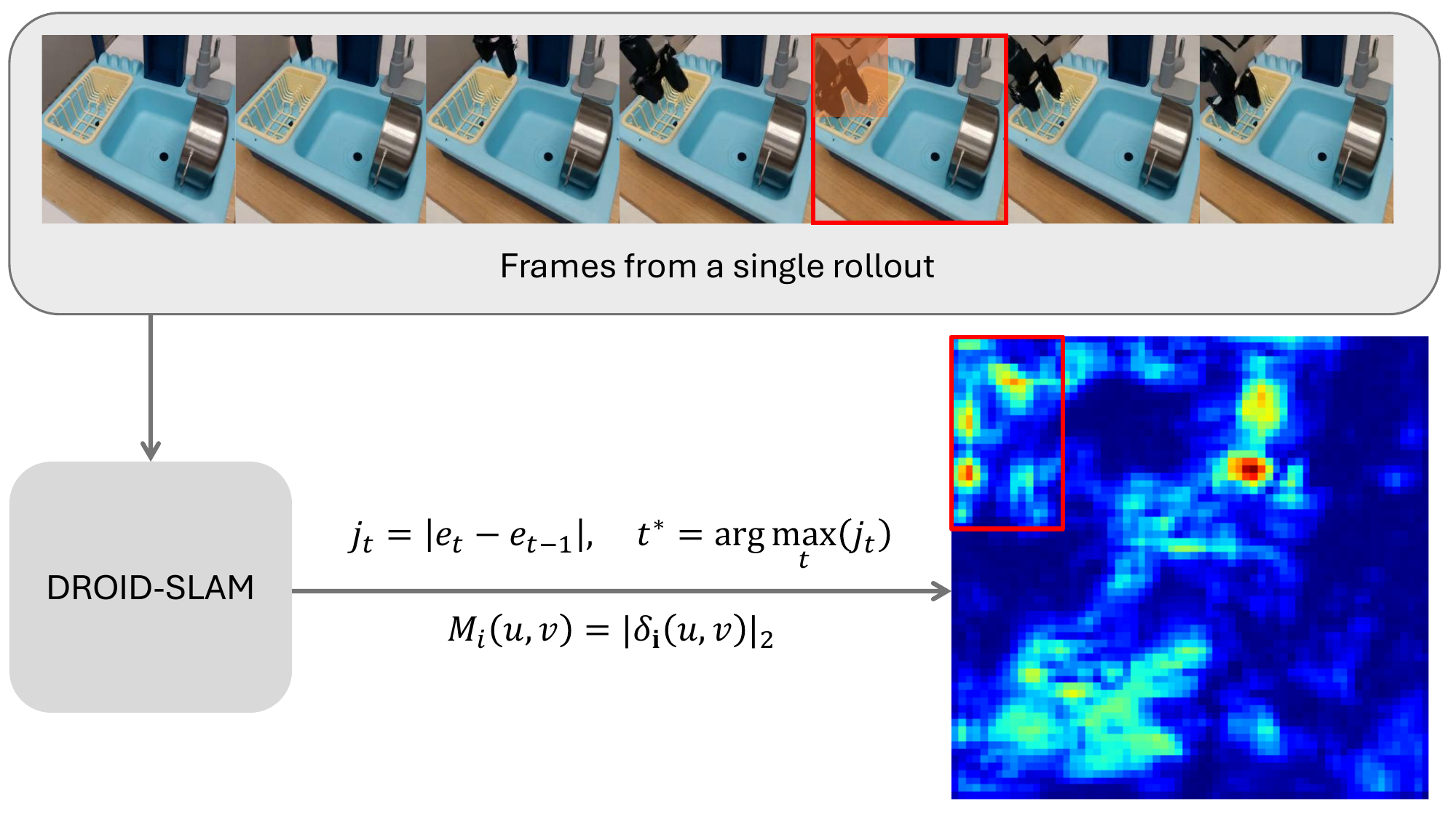}
  \caption{\textbf{Overview of Absolute Assessment}. Frames from a single rollout are evaluated using DROID-SLAM \cite{teed2022droidslamdeepvisualslam} to obtain the anomaly heatmap, which shows where and when physical feasibility collapses.}
\end{figure}

Although obvious artifacts can be easily identified, evaluating the overall quality of generated rollouts becomes challenging when multiple imperfections are present. In such cases, it is unclear which samples are more suitable for evaluation. VLA models require high temporal coherence to maintain action continuity during task execution, so our method adopts several strategies from the recent world model benchmarks \cite{duan2025worldscoreunifiedevaluationbenchmark, li2025worldmodelbenchjudgingvideogeneration}.

In this paper, we introduce relative and absolute assessments of generated rollouts: First, we introduce a relative assessment for quantitative screening. By comparing the 3D and photometric consistency of generated rollouts against real-world videos, we derive an `anomaly score’ to filter out physically implausible sequences. This pre-processing ensures that the subsequent VLM reward model evaluates only high-fidelity rollouts, preventing the policy from exploiting model artifacts. Second, we propose an absolute assessment mechanism for fine-grained qualitative diagnosis. Beyond simple filtering, we pinpoint when and where consistency collapses by employing DROID-SLAM \cite{teed2022droidslamdeepvisualslam}'s factor graphs. This generates pixel-wise heatmaps that visualize the world model's failure modes.

Overall, our work empirically improves prior methods by: (1) narrowing the focus to object morphing via a combination of entropy-based 3D analysis and photometric consistency, (2) introducing a visualization derived from geometric shift computations to point out where and when the most drastic physical failure occurs, and (3) demonstrating that our filtering mechanism improves the task success rate of VLA models by over 8\% by effectively narrowing the simulation-to-reality gap.

%% file: sec/2_method.tex
\section{Methods}
\label{sec:method}

\begin{figure*}
  \centering
  \includegraphics[width=\linewidth]{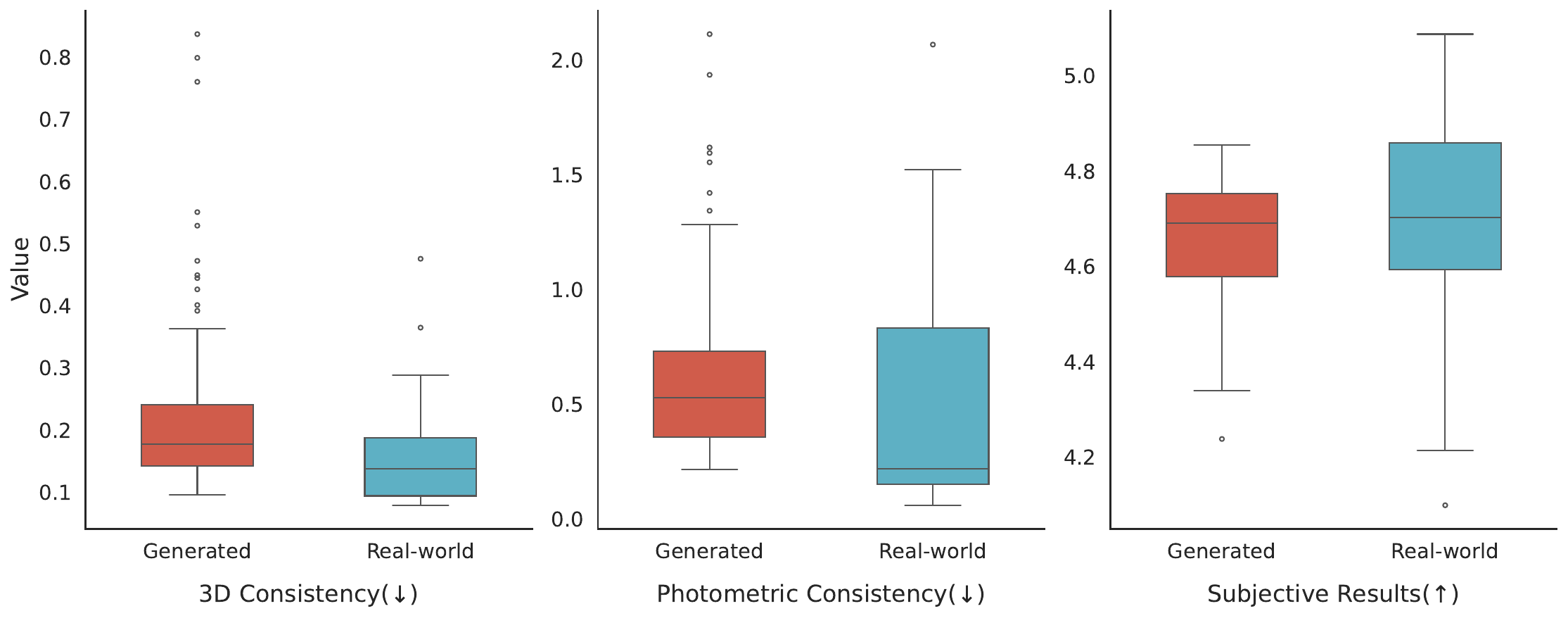}
  \caption{\textbf{Quality Scores in Real and Generated Videos}. 3D and photometric consistency exhibit distinctive gaps between real and generated distributions, whereas median of subjective quality remains comparable. Generated videos contain a significantly higher number of outliers in 3D and photometric consistency.}
  \label{fig:quality_metrics_diff}
\end{figure*}

Among the four quality metrics in WorldScore \cite{duan2025worldscoreunifiedevaluationbenchmark}, we exclude style consistency and subjective quality due to subtle variance. Then we select 3D consistency and photometric consistency, since they indicate significant statistical gaps between real and generated videos as shown in \cref{fig:quality_metrics_diff}. Accordingly, we devise relative assessment as Shannon entropy-weighted sums of consistencies, and absolute assessment based on pixel distances of adjacent frames.

\subsection{Relative Assessment: Quantitative Screening}

For 3D consistency, we calculate the reprojection error across co-visible pixels between successive frames, utilizing DROID-SLAM \cite{teed2022droidslamdeepvisualslam}. In terms of photometric consistency, we compute the Average End-Point Error(AEPE), using SEA-RAFT \cite{wang2024searaftsimpleefficientaccurate}. Both scores are computed using the publicly available source code from WorldScore \cite{duan2025worldscoreunifiedevaluationbenchmark}.

While both metrics are fundamentally significant, we utilize Shannon entropy to objectively evaluate their respective informativeness and determine the optimal weighting for each component. To evaluate the informativeness of each consistency score for anomaly detection, we calculate the Shannon entropy:

$$
H(p) = -\sum_{i=1}^{n} p(x_i) \log p(x_i)
$$

Accordingly, we calculate weights using the stable softmax function as below: $$w_k = \frac{\exp(-H_k)}{\sum_{j} \exp(-H_j)}$$
Then we assign each weight to the metrics in question.

 Subsequently, we obtain relative anomaly scores with the weighted sum:
$$
S_{\text{r}} = w_3 \cdot \hat{s}_{\text{3}} + w_p \cdot \hat{s}_{\text{p}}
$$
where $\hat{s}_k = (s_k - \mu_k) / \sigma_k$ denotes the z-score normalized score of the metric $k$. The stability-based weighting mechanism ensures that the anomaly detection is robust even when one metric exhibits high variance due to generation noise.

\subsection{Absolute Assessment: Qualitative Analysis}

When there are not enough samples, we can borrow relative anomaly scores from real-world recordings in similar environments. However, since the score alone is not sufficient to evaluate overall quality, we propose a two-step analysis based on DROID-SLAM \cite{teed2022droidslamdeepvisualslam} factor graph, focusing on 3D consistency to identify specific failure points.

We specifically prioritize 3D consistency for absolute assessment and qualitative visualization. This choice is justified by its lower Shannon entropy compared to photometric consistency, indicating higher informational certainty. Furthermore, the reprojection error in 3D consistency provides a more direct physical grounding, enabling us to precisely localize where and when physical anomalies occur within the generated sequence.

\subsubsection{Temporal Detection} 
We first monitor the reprojection error $e_t$ from the factor graph updates. To exclude the initial warmup, we only consider frames where $t > 7$, following the original implementation. The frame $t^*$ with the maximum geometric shift is then identified by the error jump $j_t$:
$$
j_t = |e_t - e_{t-1}|, \quad t^* = \arg\max_{t} (j_t)
$$

\subsubsection{Spatial Localization} 
Within the selected frame $t^*$, we examine the residual flow $\mathbf{\delta}_i \in \mathbb{R}^{H \times W \times 2}$ for each edge $i$ on the factor graph, where $H$ and $W$ denote the height and width of the frame. Since each $\mathbf{\delta}_i$ is a dense vector field which contains error values for all pixels, we calculate the pixel-wise magnitude $M_i(u, v)$ to visualize the intensity of distortions:
$$
M_i(u, v) = \|\mathbf{\delta}_i(u, v)\|_2
$$
where $(u, v)$ represents the coordinates of the pixels.

\begin{table*}
\centering
\begin{tabular}{llccc}
\toprule
\textbf{Category} & \textbf{Task} & \textbf{\# Trials} & \makecell{\textbf{WorldGym} \\ \textbf{\# Successes}} &
\makecell{\textbf{Real-world} \\ \textbf{\# Successes}} \\
\midrule
Visual gen & Put Eggplant into Pot (Easy Version) & 10 & 7   & 10 \\
Motion gen & Lift Eggplant         & 10 & 7.5 & 7.5 \\
Physical gen & Put Carrot On Plate   & 10 & 4   & 8 \\
Semantic gen & Stack Blue Cup on Pink Cup & 10 & 6 & 4.5 \\
Language Grounding & Put \{Blue Cup, Pink Cup\} on Plate & 10 & 8.5 & 9.5 \\
\midrule
& & Mean Success Rate & \textbf{66.0} $\pm$ 6.6\% & \textbf{79.0} $\pm$ 5.5\% \\
\bottomrule
\end{tabular}
\caption{\textbf{Baseline Task Performance from Prior Work}. Task success numbers are taken directly from WorldGym \cite{quevedo2025worldgymworldmodelenvironment}. Average success rates $\pm$ StdErr are computed across 50 total rollouts, following OpenVLA calculated across 170 total rollouts for 17 tasks.}
\label{tab:task_results}
\end{table*}

To select the most representative artifact, we calculate the mean magnitude $S_i$ for each edge:
$$
S_i = \frac{1}{HW} \sum_{u,v} M_i(u, v), \quad i^* = \arg\max_{i} (S_i)
$$

%% file: sec/3_exp.tex
\section{Experiments}
\label{sec:experiments}

We specifically focus our evaluation on OpenVLA \cite{kim2024openvlaopensourcevisionlanguageactionmodel} due to the comprehensive accessibility of real-world out-of-distribution(OOD) of BridgeData V2 \cite{walke2024bridgedatav2datasetrobot} with WidowX robot experiment videos and success rates. While our methodology applies to any robot policy evaluation, this allows a quantitative validation of our proposed measures by comparing generated low-anomaly rollouts against established real-world performance.

5 tasks are selected based on the availability of real-world experiment videos on the OpenVLA website, with one task representing each of the categories suggested in their study. The details of the five tasks are summarized in \cref{tab:task_results}. Then 30 videos per task are generated employing WorldGym \cite{quevedo2025worldgymworldmodelenvironment} with a random seed and OpenVLA.

\subsection{Relative Assessment}

To sort high and low anomaly rollouts, we generate 30 videos for each of the five tasks, resulting in a total of 150 videos. To ensure variability, each video was produced using a unique random seed, starting from 42 and incrementing for each subsequent generation.

Both real-world and generated videos' scores are put into equal-width bins(n=30), where $p(x_i)$ denotes the frequency of scores falling into the $i$-th bin. The number of bins was chosen to align with the sample size of our real-world dataset. Analysis across these datasets shows that 3D consistency has lower entropy compared to photometric consistency, which can be interpreted as 3D consistency having higher informational certainty and representational stability. Specific numbers are described in \cref{tab:entropy_weights}.

\begin{table}[ht]
\centering
\begin{tabular}{llcc}
\toprule
Metric & Property & Generated & Real-world \\
\midrule
\multirow{2}{*}{3D Consist} & Entropy & 2.16 & 1.73 \\
& Weight  & \textbf{0.67} & \textbf{0.65} \\
\midrule
\multirow{2}{*}{Photo Consist} & Entropy & 2.88 & 2.35 \\
& Weight  & 0.33 & 0.35 \\
\bottomrule
\end{tabular}
\caption{\textbf{Shannon Entropy and Weights of Real and Generated Videos}. The weights derived from generated and real-world datasets align, even with disparity between two datasets.}
\label{tab:entropy_weights}
\end{table}

\begin{figure}[ht]
  \centering
  \includegraphics[width=\linewidth]{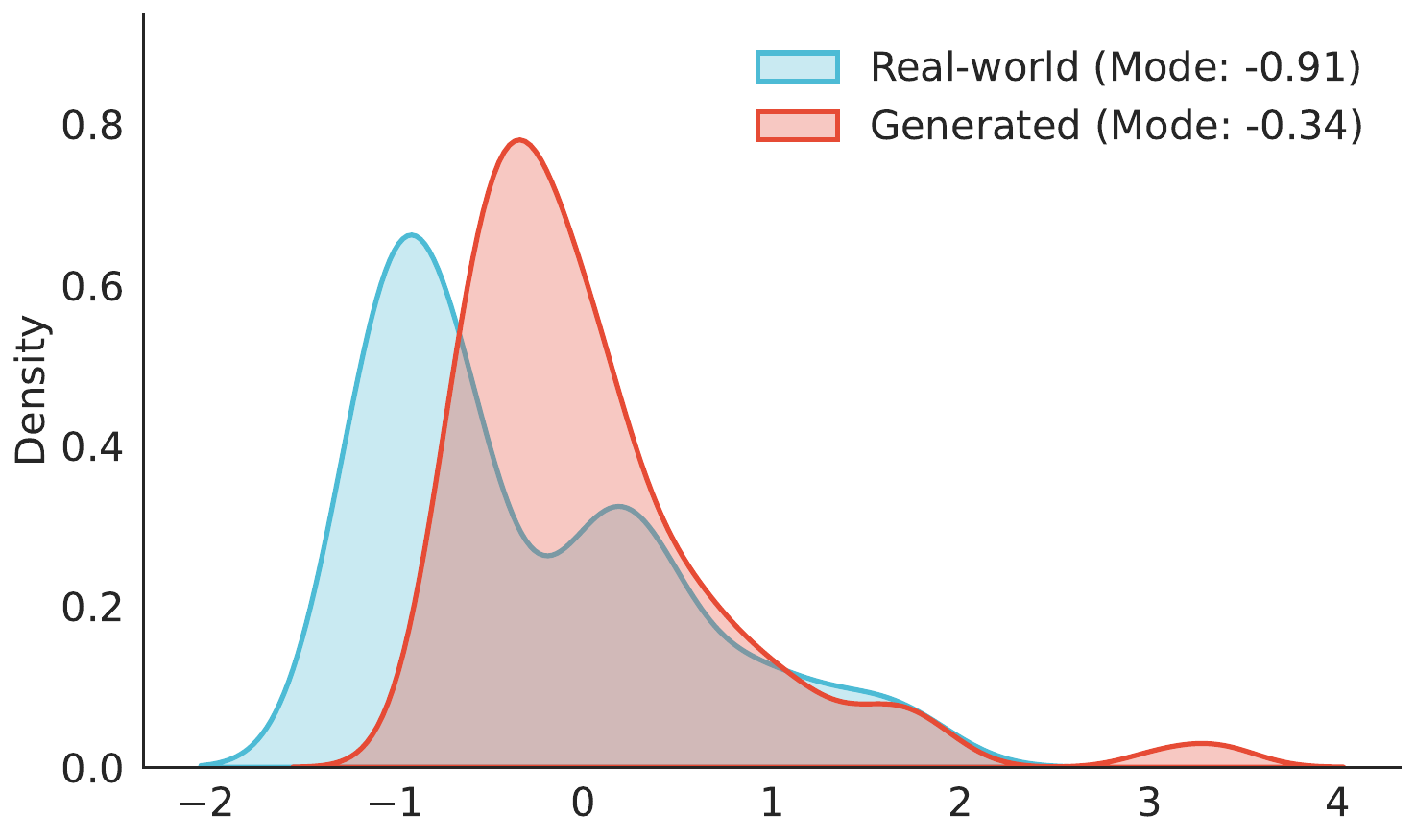}
  \caption{\textbf{Distribution of Relative Anomaly Scores}. Lower score is better. Density allows normalized comparison for varied OpenVLA rollouts across unequal sizes.}
  \label{fig:relative_anomaly_pdf}
\end{figure}

The distribution of relative anomaly scores \cref{fig:relative_anomaly_pdf} shows a noticeable shift, where generated videos tend to have higher scores and wider variance compared to real-world videos. This suggests that generated videos with low anomaly scores may better represent the real-world capabilities of VLA models.

\begin{figure}[ht]
  \centering
  \includegraphics[width=\linewidth]{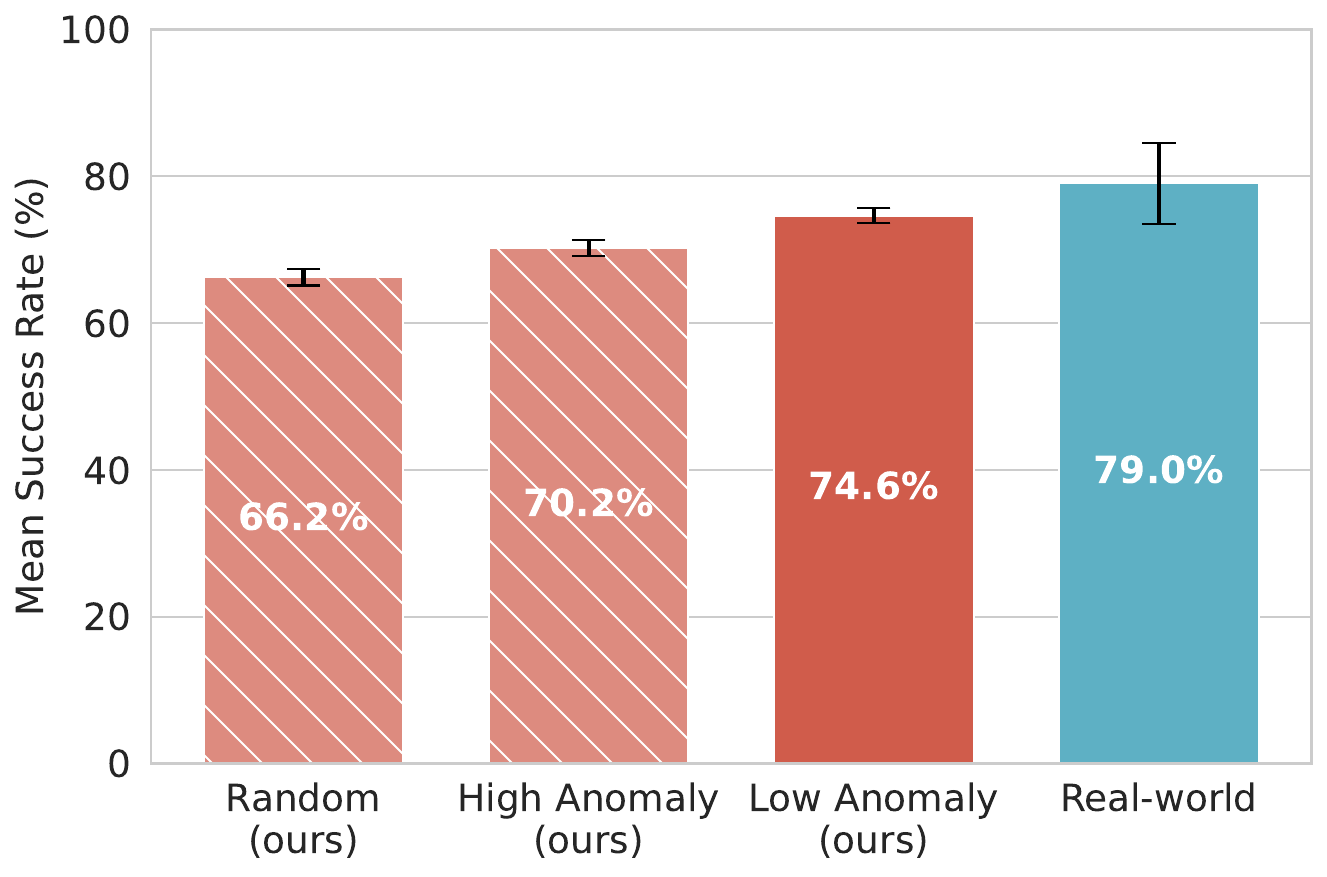}
  \caption{\textbf{Success Rates across different conditions} after filtering with our relative assessment with the real-world result \cite{kim2024openvlaopensourcevisionlanguageactionmodel}. Our results show that the low-anomaly group achieves success rates closest to those of real-world videos. We repeat GPT-4o evaluation 10 times per video to reduce evaluator variance, resulting in a smaller SE compared to real-robot experiments. Detailed per-task results are provided in \cref{tab:rel_score_results}}
  \label{fig:eval_by_condition}
\end{figure}

To evaluate the generated rollouts, GPT-4o \cite{openai2024gpt4ocard} was employed as a reward model. The evaluation follows the approach described in WorldGym \cite{quevedo2025worldgymworldmodelenvironment}. Rollouts per task were categorized into high-anomaly and low-anomaly groups (n=10 each) based on their relative assessment scores, while excluding the middle 10 rollouts to ensure better distinction between the two groups. Additionally, as a baseline for comparison, we evaluate 10 randomly generated rollouts without fixed seeds or anomaly-based filtering. The result is described in \cref{fig:eval_by_condition}

\begin{table*}
\centering
\begin{tabular}{lccccc}
\toprule
Task &
\makecell{Random \\ (ours)} &
\makecell{High Anomaly \\ (ours)} &
\makecell{Low Anomaly \\ (ours)} &
Real-world \cite{kim2024openvlaopensourcevisionlanguageactionmodel} \\
\midrule
Put Eggplant into Pot (Easy Version) & 70 & 86 & 73 & 100 \\
Lift Eggplant        & 74 & 84 & 64 & 75 \\
Put Carrot On Plate  & 48 & 39 & 75 & 80 \\
Stack Blue Cup on Pink Cup & 63 & 72 & 79 & 45 \\
Put \{Blue Cup, Pink Cup\} on Plate & 76 & 70 & 82 & 95 \\
\midrule
Mean Success Rate & \textbf{66.2} $\pm$ 1.1\% & 
\textbf{70.2} $\pm$ 1.1\% &
\cellcolor{gray!15} \textbf{74.6} $\pm$ 1.0\% &
\cellcolor{gray!15} \textbf{79.0} $\pm$ 5.5\% \\
\bottomrule
\end{tabular}
\caption{\textbf{Task Performance (\%)} shows relative assessment results. We select 5 tasks in 17 OOD tasks from OpenVLA \cite{kim2024openvlaopensourcevisionlanguageactionmodel}. Each task is selected to follow five categories as we elaborate earlier \cref{sec:experiments}. Random, High Anomaly, Low Anomaly group data is from our experiments, which conduct 10 sets of 10 trials (100 iterations) per task. Real-world data is sourced from OpenVLA \cite{kim2024openvlaopensourcevisionlanguageactionmodel} as well as \cref{tab:task_results}, which reports results based on 10 trials per task. Among three groups of generated rollouts, Low Anomaly group by our metric yield a mean success rate that approximates Real-world performance.}
\label{tab:rel_score_results}
\end{table*}

Applying relative anomaly scores for filtering improves evaluation accuracy and helps bridge the simulation-to-reality gap. However, we observe a specific limitation in tasks like `Lift Eggplant', where low-anomaly rollouts fail to capture the real-world VLA ability in \cref{tab:rel_score_results}. This is because when the robot fails to contact the object entirely, the lack of visual distortion results in misleadingly low scores. This confirms that physical consistency is a necessary but insufficient condition for semantic success, which we leave for future studies.

\subsection{Absolute Assessment}

The rollouts used are initialized from a frame where the robot arm is outside the field of view. Consequently, the world model lacks sufficient visual grounding to accurately reconstruct the arm's morphology, leading to inconsistent or hallucinated representations in the generated sequence.

The resulting magnitude map $M_{i^*}$ is used as our Artifact Heatmap. This highlights specific regions where the generative model fails to maintain 3D consistency. 

\begin{figure} [ht]
    \centering
    \begin{subfigure}[b]{0.48\columnwidth}
        \centering
        \includegraphics[width=\textwidth]{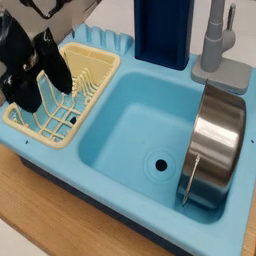}
    \end{subfigure}
    \hfill
    \begin{subfigure}[b]{0.48\columnwidth}
        \centering
        \includegraphics[width=\textwidth]{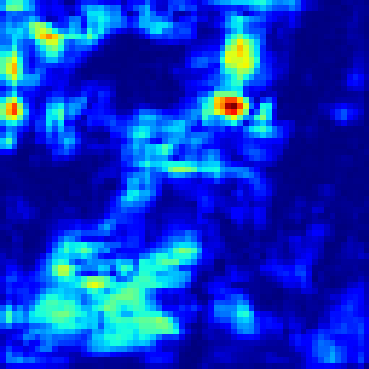}
    \end{subfigure}

    \vspace{5pt}

    \begin{subfigure}[b]{0.48\columnwidth}
        \centering
        \includegraphics[width=\textwidth]{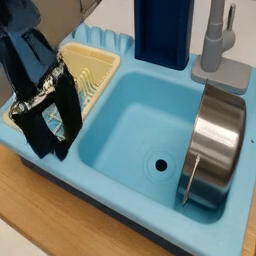}
    \end{subfigure}
    \hfill
    \begin{subfigure}[b]{0.48\columnwidth}
        \centering
        \includegraphics[width=\textwidth]{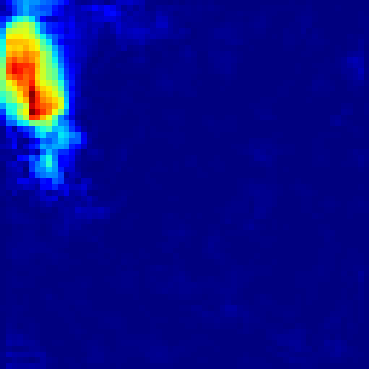}
    \end{subfigure}

    \caption{\textbf{Artifact Heatmap} shows the result of the absolute assessment. Red colors imply low fidelity of the specific edges. The first row displays a deformed robot arm and the second row shows a thickened robot arm, both of which are clearly characterized in the adjacent heatmap.}
    \label{fig:artifact_heatmap}
\end{figure}

\subsection{Challenges in Cross-Model Generalization}

To stress-test the robustness of this metric across diverse distributions, we extended our evaluation to generated clips from representative models \cite{sora, grok, decart2025lucyedit}.

\textbf{Sora}. The clip utilized in our analysis is the 'woman walking down a Tokyo street' sourced from the official OpenAI website. While the video exhibits synthetic artifacts easily recognizable by humans, these anomalies remain consistent throughout the sequence. Consequently, the 3D consistency score does not exhibit significant fluctuations, rendering the analysis of a single captured frame unreliable. Indeed, the sequence recorded a Mean Score of 1.9326 with a Max Jump of only 0.0063, confirming the high temporal persistence of artifacts.

\begin{figure}[ht]
  \centering
  \includegraphics[width=\linewidth]{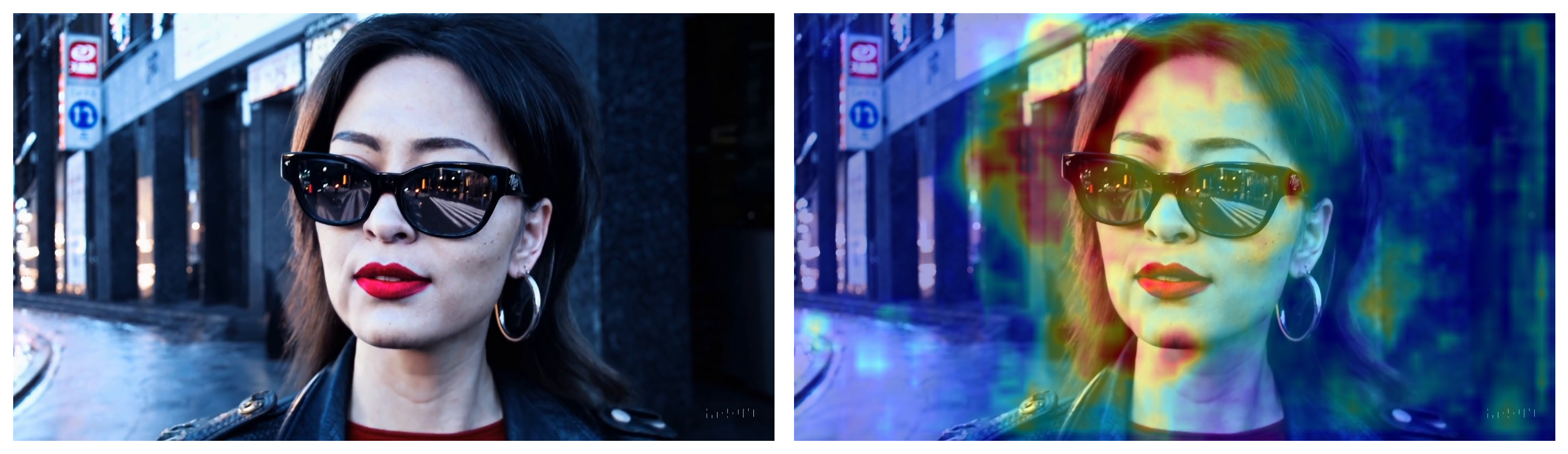}
  \caption{\textbf{Sora}. The left image shows the original frame, and the heatmap is overlaid on the original image. This is not a relatively implausible frame to human perspective.}
\end{figure}

\textbf{Grok}. We design a prompt describing a hand grasping an apple that exhibits surreal, gelatinous deformation instead of rigid-body physics. In this case, the frame in question was detected, but the heatmap was distributed broadly over the sequence, lacking the spatial precision to isolate the point of consistency collapse. The sequence recorded a Mean Score of 1.6282 with a Max Jump of 0.0041.

\begin{figure}[ht]
  \centering
  \includegraphics[width=\linewidth]{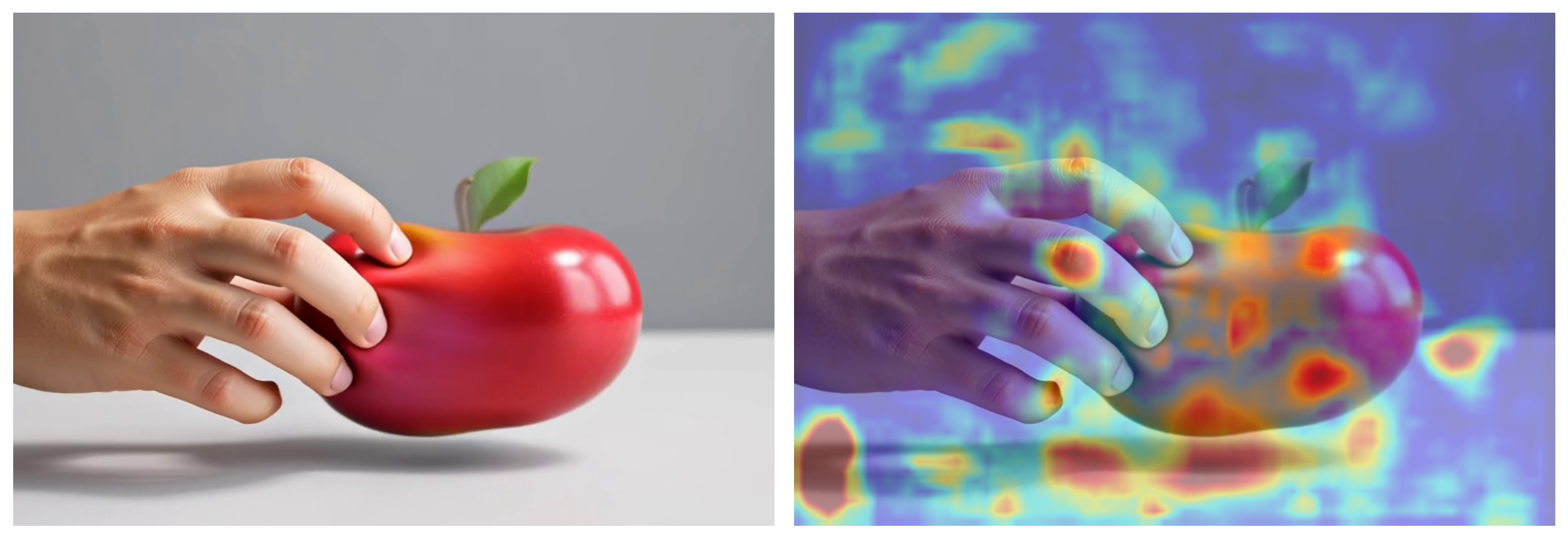}
  \caption{\textbf{Grok-generated physically implausible sequence and the heatmap}. The left image shows the original frame, and the heatmap is overlaid on the original image. The heatmap shows the limitation in our method to capture specific location where the physical inconsistency is detected in the frame.}
\end{figure}

\textbf{LucyEdit}. Because LucyEdit often produces artifacts from the frame, our detector paradoxically identifies the `improved' frame as an anomaly. Although it successfully captures a deviation in the sequence, it fails to isolate physical failure due to the absence of a clean reference frame. The input natural video recorded a mean score of 0.1874 with a max jump of 0.0039 and the output edited footage showed a mean score of 0.1432 with a max jump of 0.0038, which demonstrates certain generated footages are more consistent than natural videos according to the reprojection error.

\begin{figure}[t]
    \centering
    \begin{subfigure}[b]{0.48\columnwidth}
        \centering
        \includegraphics[width=\textwidth]{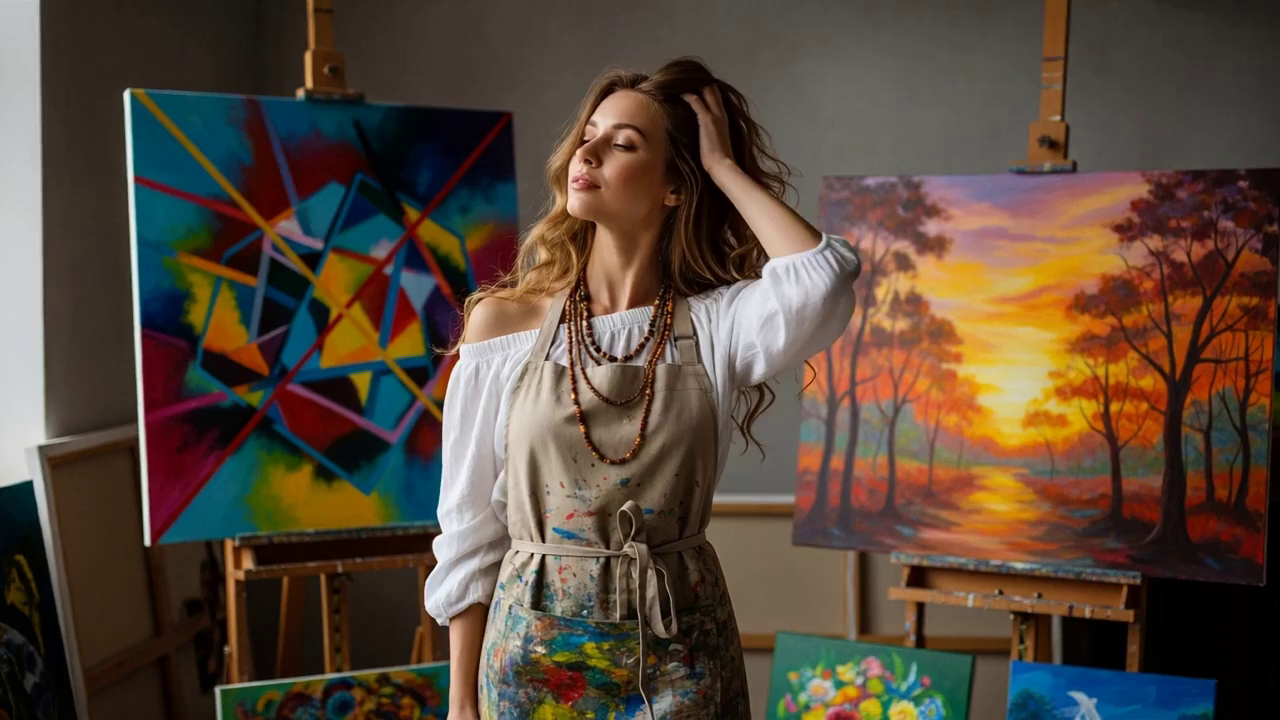}
    \end{subfigure}
    \hfill
    \begin{subfigure}[b]{0.48\columnwidth}
        \centering
        \includegraphics[width=\textwidth]{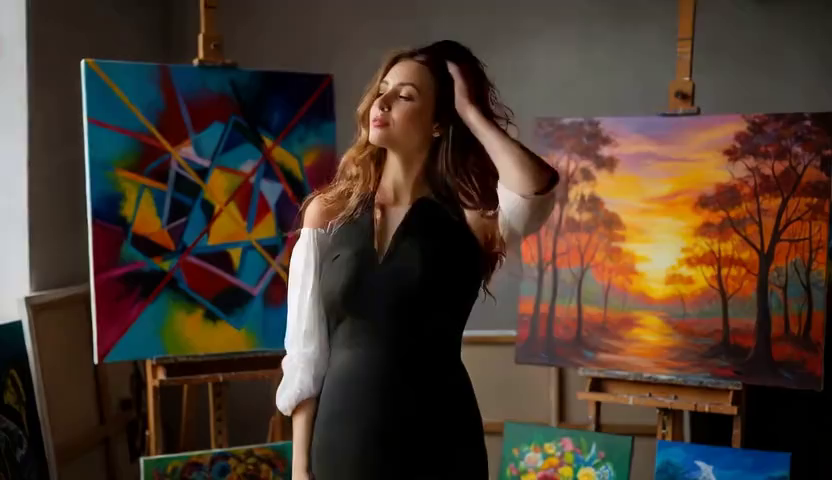}
    \end{subfigure}
    \caption{\textbf{LucyEdit} original sample's first frame (left) and the edited footage's first frame (right). In the edited frame, a distinct 'V' shape artifact is visible on her chest around the black garment. This contour is unnatural and does not correspond to either clothing drape or anatomy.}
\end{figure}

\begin{figure}[ht]
  \centering
  \includegraphics[width=\linewidth]{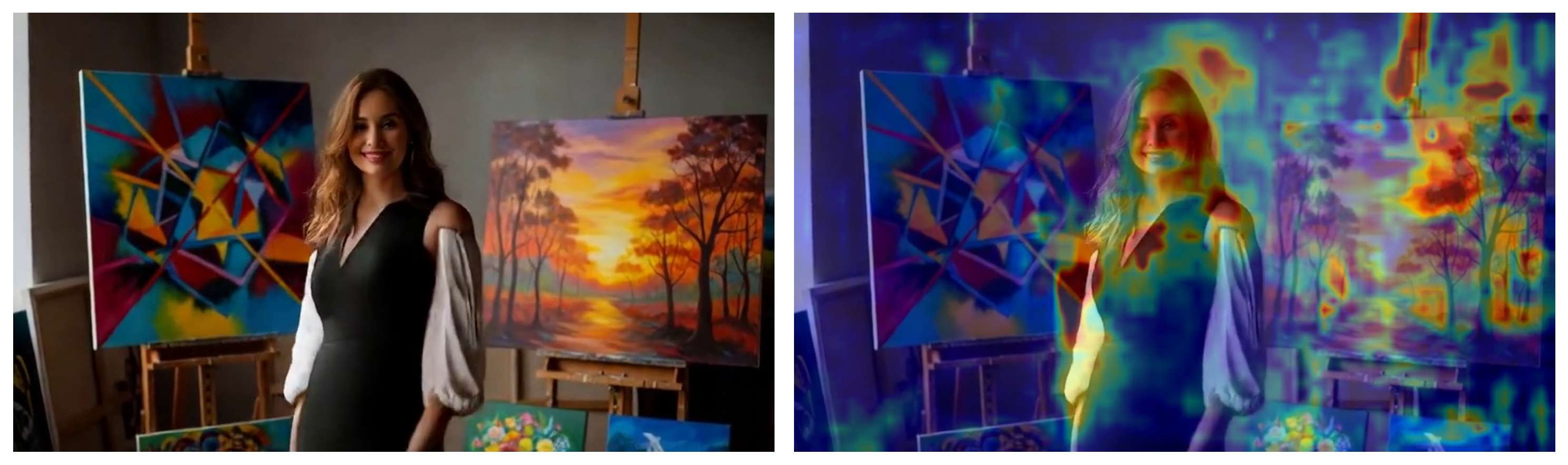}
  \caption{The edited sequence from \textbf{LucyEdit} displays more significant artifacts in the preceding frames compared to this sample. The specific editing instruction provided to the model was: `Change her clothes to a black mini dress for funeral'.}
\end{figure}

\textbf{Noise-corrupted videos}. Stable Diffusion 3 \cite{esser2024scalingrectifiedflowtransformers} is chosen to create an anomaly frame. We use a custom recorded footage, and we also use generated videos from the three models mentioned above.

\begin{figure}
    \centering
    \begin{subfigure}[b]{0.48\columnwidth}
        \centering
        \includegraphics[width=\textwidth]{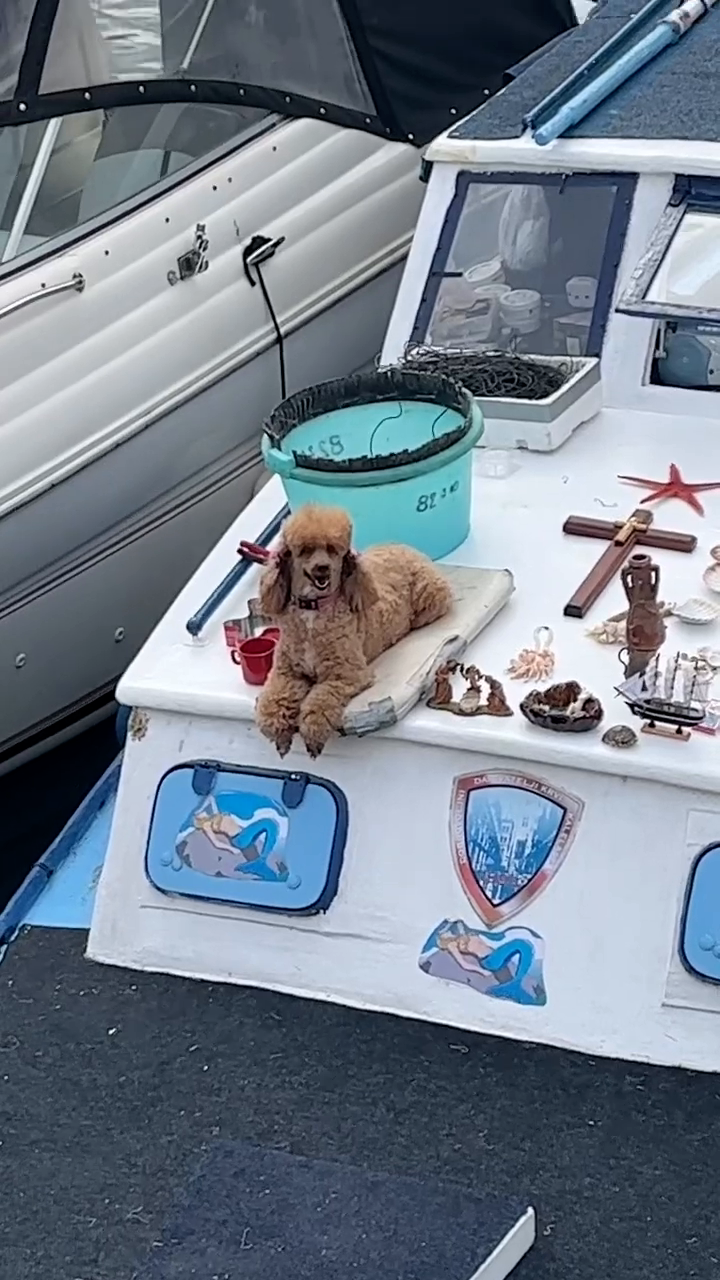}
    \end{subfigure}
    \hfill
    \begin{subfigure}[b]{0.48\columnwidth}
        \centering
        \includegraphics[width=\textwidth]{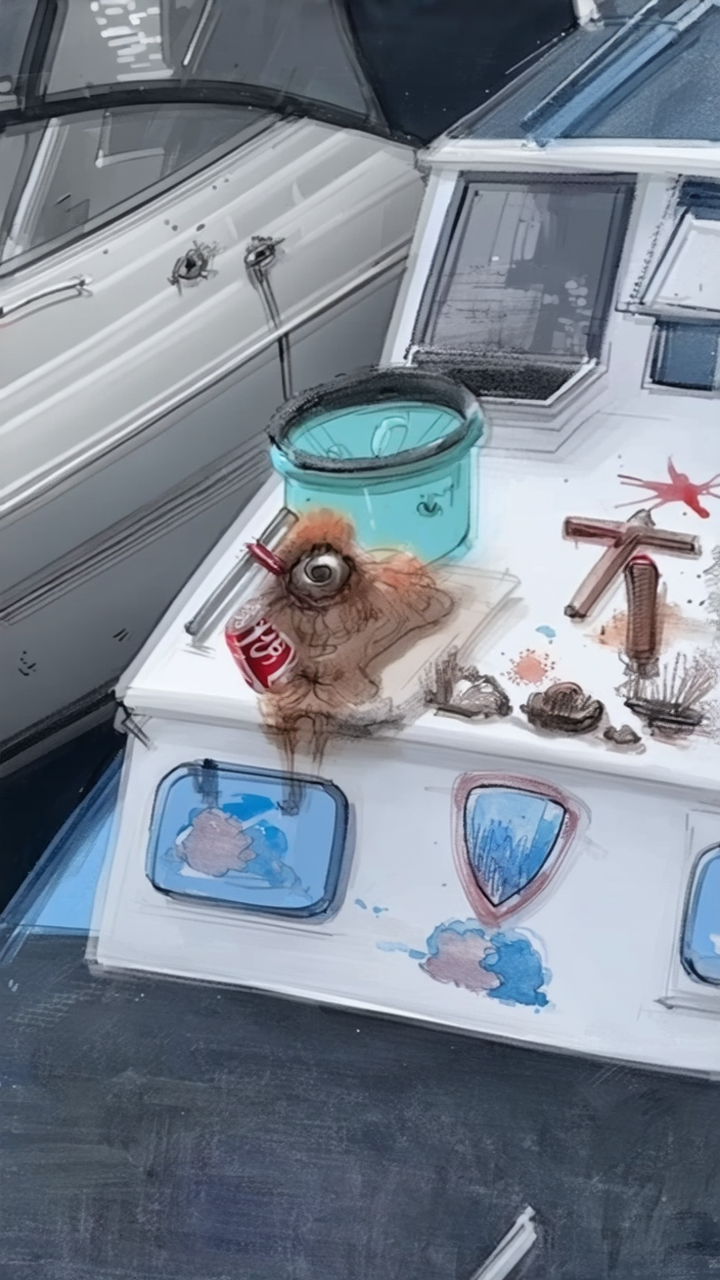}
    \end{subfigure}
    \caption{\textbf{Noise-corrupted}. These are one of samples, which is a custom recorded footage. The left one is the original frame, and the right one is the corrupted image that replaced the original frame for the experiments. Both the heatmap generation and score computation fail due to zero values encountered during the calculation.}
\end{figure}

With several versions of augmentation applying various prompts and hyperparameter tuning, our method always returns undesired frames that are not corrupted. In sequences containing augmented frames, the reprojection error occasionally drops to zero, which likely indicates a failure to establish valid pixel correspondences. Standard video sequences, whether generated or natural, consistently yield non-zero scores, reflecting a continuous flow of identifiable features.

This cross-model analysis identifies two fundamental challenges for current SLAM-based assessment: (1) the collapse of tracking under severe structural inconsistencies where correspondences cannot be established, and (2) localization instability under non-rigid deformations or dynamic lighting. Addressing these challenges remains an important direction for future work.

%% file: sec/4_discussion.tex
\section{Discussion}
\label{sec:discussion}

\textbf{Conclusion}. We present a practical application of existing consistency metrics for analyzing generated sequences, specifically within the robotic simulation domain. This approach does not require a ground truth, which is often unavailable when assessing outputs from modern generative models. In scenarios where ground-truth references are unavailable, our metrics serve as a preliminary filter to ensure basic physical consistency. By narrowing the simulation-to-reality gap in VLA model evaluation, this approach offers a scalable alternative to manual inspection for verifying robot-centric visual rollouts. By extension, our relative anomaly score can serve as a reward signal for refining world models.

\noindent \textbf{Limitation}. Our assessment relies on quality metrics devised for static generation, which can fail under severe structural inconsistencies, producing zero values that limit applicability to certain generation models. Anomaly scores can also be misleadingly low when task failure involves no physical contact. Beyond these metric-level limitations, our evaluation is primarily conducted on OpenVLA, leaving generalization to other VLA models unexplored. Future work will extend evaluation across diverse VLA and world models, and investigate semantic or contact-aware signals to address these limitations.

\section*{Acknowledgement}
We thank the authors of WorldGym and WorldScore for providing the open-source codebases that served as the foundation for our experiments. We are also grateful to the OpenVLA contributors for their real-world experiment rollouts, which were essential for the execution of our comparative analysis.